\documentclass[letterpaper, 10 pt, conference]{ieeeconf}
\IEEEoverridecommandlockouts                              
\usepackage{comment}

\usepackage{algorithm}
\usepackage{algorithmic}
\usepackage{url}
\usepackage{amsmath}
\usepackage{cite}
\usepackage[dvips]{graphicx}
\usepackage{epsfig}
\usepackage{epic,eepic,color}
\usepackage{times}
\usepackage{mathptmx}
\usepackage{multirow}
\usepackage{stkernel}

\usepackage{cases}

\usepackage{times}
\usepackage{multirow}

\definecolor{red}{rgb}{1,0,0}
\definecolor{green}{rgb}{0,1,0}
\definecolor{blue}{rgb}{0,0,1}
\definecolor{violet}{rgb}{1,0,1}
\definecolor{cyan}{cmyk}{1,0,0,0}
\definecolor{magenta}{cmyk}{0,1,0,0}
\definecolor{yellow}{cmyk}{0,0,1,0}

\definecolor{white}{rgb}{1,1,1}

\usepackage{jumoline}

\newcommand{\CO}[1]{}

\newcommand{\CommentOut}[1]{}

 \newcommand{\editage}[1]{}

\newcommand{\FIG}[3]{
\begin{minipage}[b]{#1cm}
\begin{center}
\includegraphics[width=#1cm]{#2}\\
{\scriptsize #3}
\end{center}
\end{minipage}
}

\newcommand{\FIGR}[3]{
\begin{minipage}[b]{#1cm}
\begin{center}
\includegraphics[angle=-90,clip,width=#1cm]{#2}
\\
{\scriptsize #3}
\vspace*{1mm}
\end{center}
\end{minipage}
}

\newcommand{\figA}{
\begin{figure}[t]
\vspace*{-2mm}
  \begin{center}
\FIG{8}{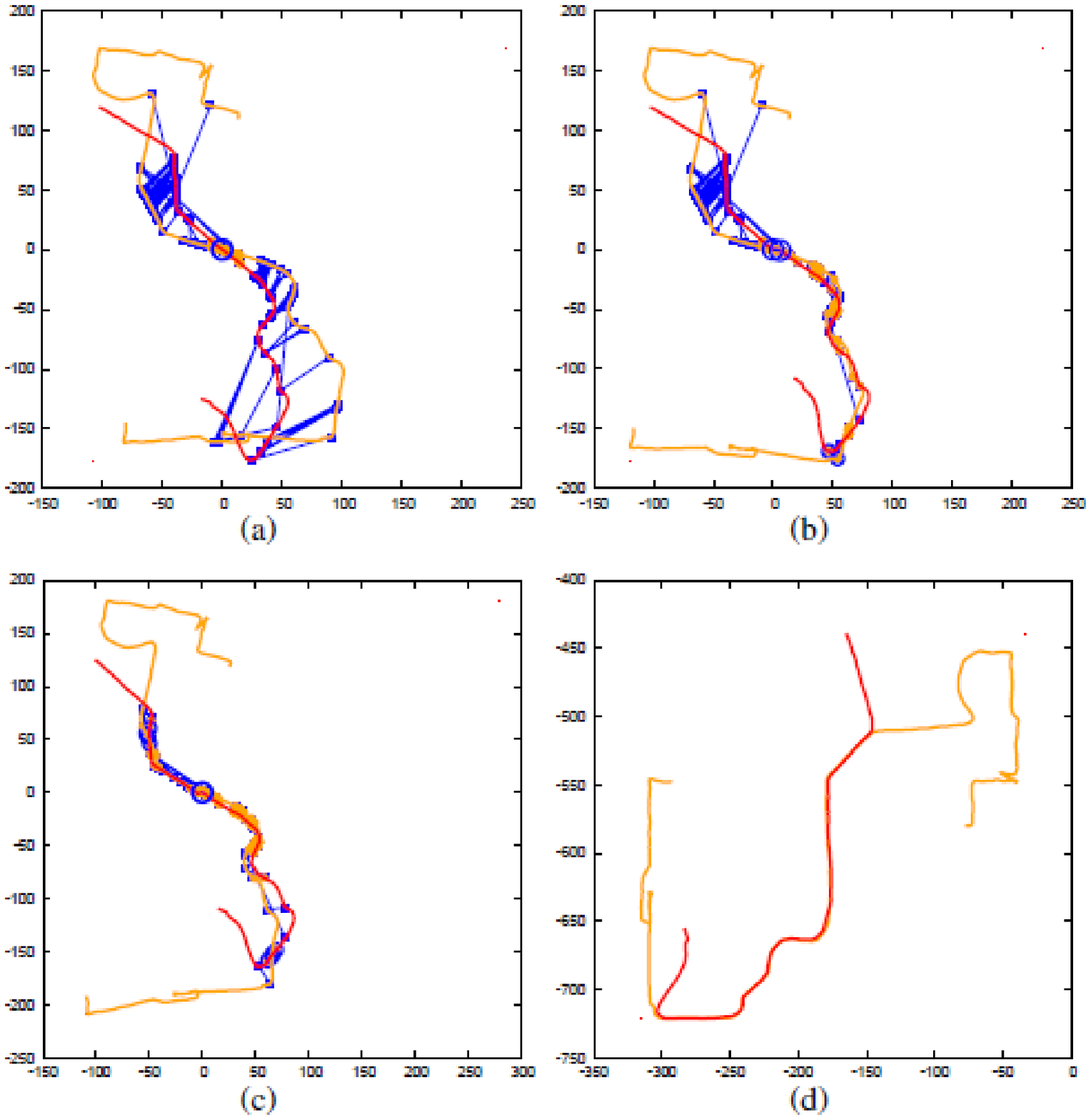}{}
\caption{
Deformable map matching for uncertain loop-less maps. Although neither the query (orange curved line) nor the reference map (red curved line) has a large loop, many loop closure constraints (blue straight line) are found by merging the two maps. The big blue circles indicate an image correspondence at which the two maps are merged, while the small blue circles indicate additional loop closure constraints used by the proposed algorithm. (a) Previous map-matching methods with the assumption of rigid transformation. (b, c) Proposed deformable map-matching method with single loop and with multiple loops, respectively. (d) Ground-truth map-matching result. In all the figures, datasets ``01" and ``03" are used as query and reference. 
}\label{fig:A}
\end{center}
\vspace*{-7mm}
\end{figure}
}

\newcommand{\figBb}[3]{
  \begin{minipage}[b]{6.2cm}
  \begin{center}
\hspace*{-15mm}\FIGR{4}{figs/fig2a_dir/#1_1_0.eps}{}\hspace*{-1.5cm}%
\FIGR{4}{figs/fig2a_dir/#1_1_2.eps}{}\hspace*{-15mm}\vspace*{1mm}\\
\hspace*{-15mm}\FIGR{4}{figs/fig2a_dir/#1_1_10000.eps}{}\hspace*{-1.5cm}%
\FIGR{4}{figs/fig2a_dir/#1_1_gt.eps}{}\hspace*{-15mm}\vspace*{-3mm}\\
{\scriptsize query:#2, ref:#3}
  \end{center}
  \end{minipage}
  \hspace*{-7mm}
}

\newcommand{\figBc}[1]{
\figBb{20120122_20120804}{01}{08}\figBb{20120122_20121117}{01}{11}\figBb{20120331_20120122}{03}{01}\\
\figBb{20120331_20120804}{03}{08}\figBb{20120331_20121117}{03}{11}\figBb{20120804_20120122}{08}{01}\\
\figBb{20121117_20120122}{11}{01}\figBb{20121117_20120331}{11}{03}\figBb{20121117_20120804}{11}{08}\\
}

\newcommand{\figB}{
\begin{figure*}[t]
  \begin{center}
\FIG{17}{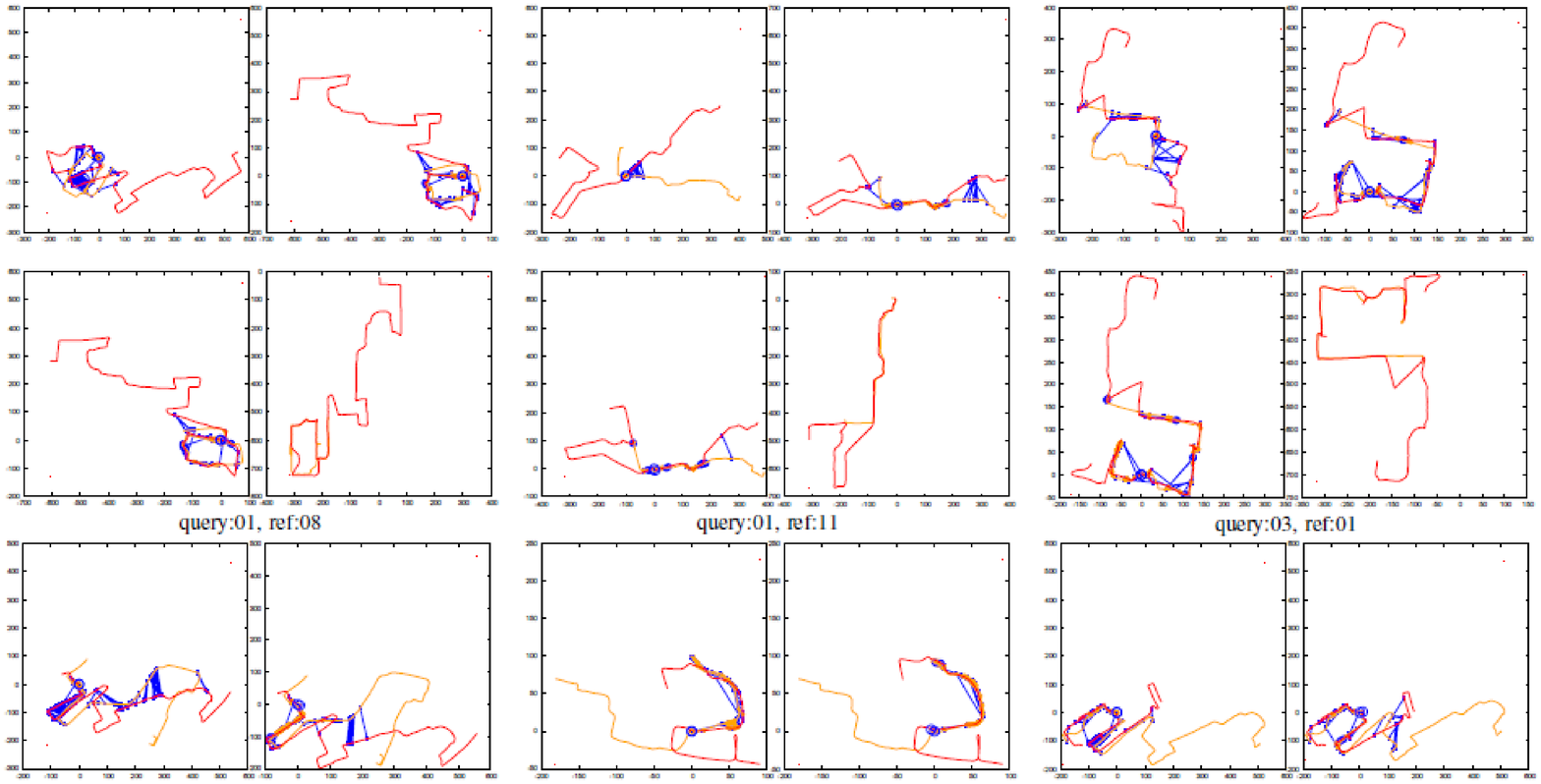}{}\\
\FIG{17}{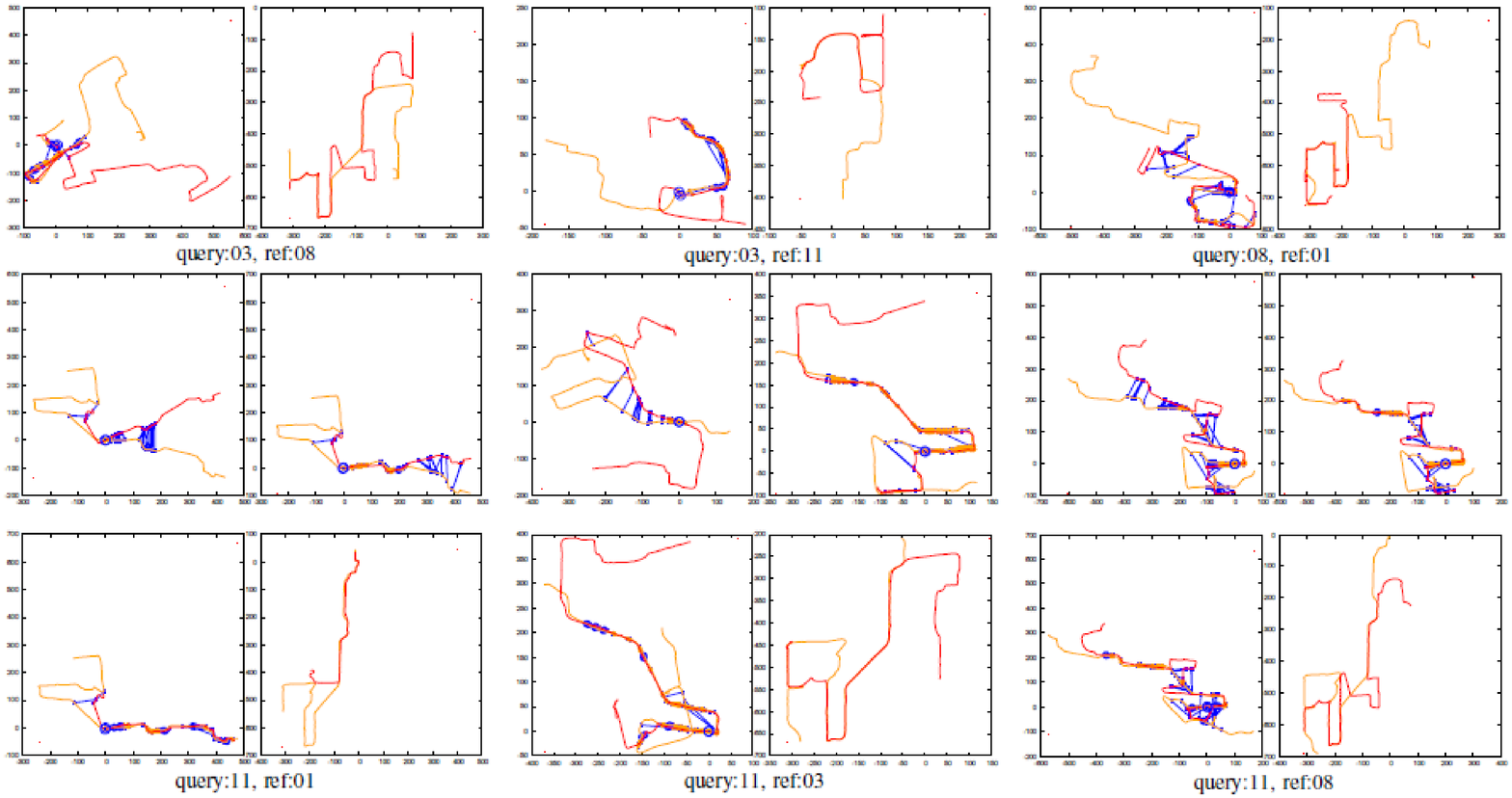}{}\\
\caption{
Example map-matching results. Shown are map-matching samples from nine out of the 10 combinations of query and reference datasets, except for the one combination shown in Fig. \ref{fig:A}. In each example, the top left, top right, and bottom left panels show the results of map matching by the ``naive," the ``single," and the ``multiple" algorithm, respectively, while the bottom right panel shows the ground-truth result. The meanings of the orange, red, and blue lines and blue circles remain the same as in Fig.\ref{fig:A}.
}\label{fig:B}
\end{center}
\vspace*{-5mm}
\end{figure*}
}

\newcommand{\figC}{
\begin{figure}[t]
  \begin{center}
\FIG{8}{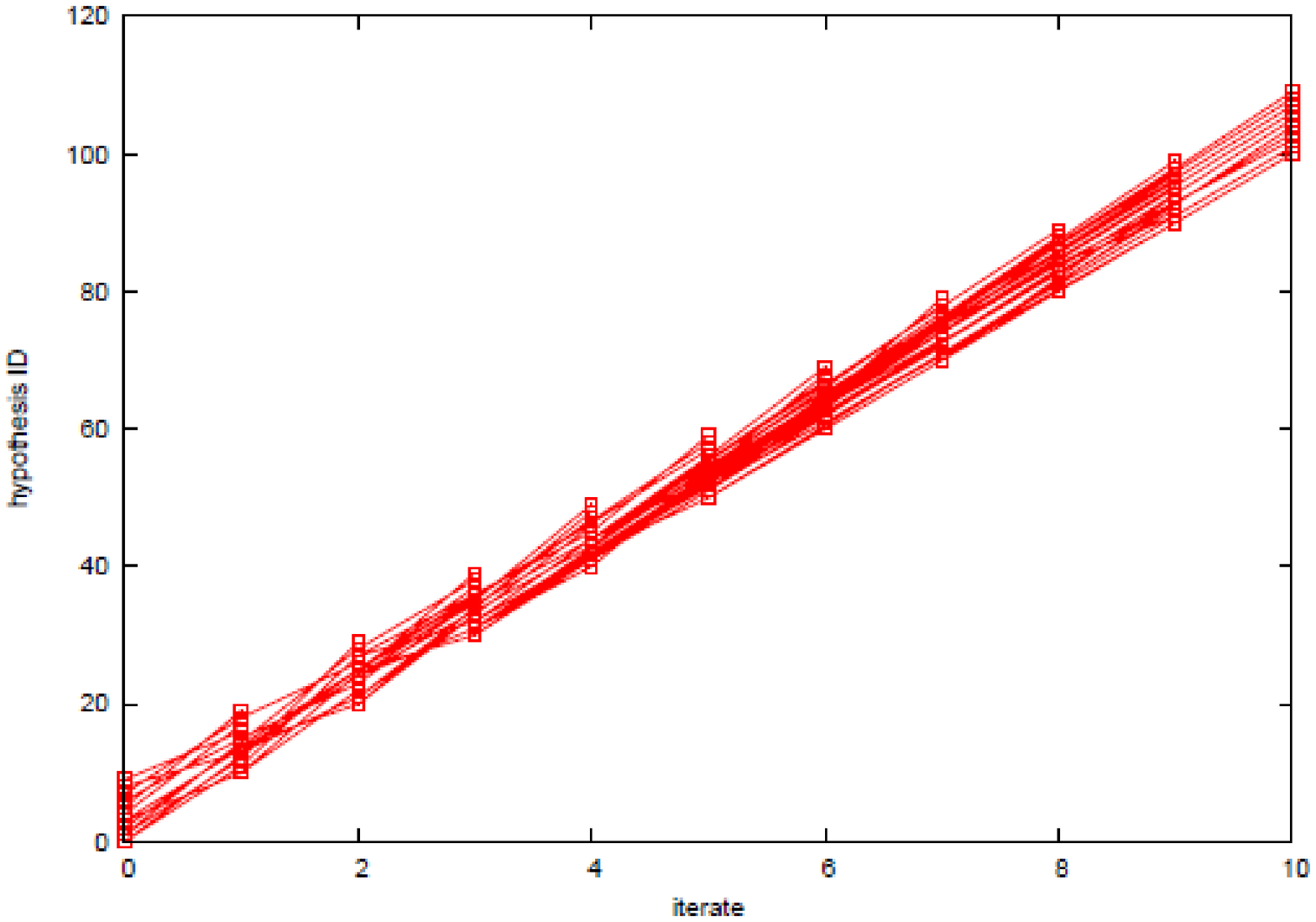}{}
\caption{
Sequential hypothesis generation. Horizontal axis: iteration ID; Vertical axis: hypothesis ID. Each line connects one of the $K'$ hypotheses top-ranked at each iteration to the next-generation hypothesis generated from it.
}\label{fig:C}
\end{center}
\end{figure}
}

\newcommand{\figD}{
\begin{figure}[t]
  \begin{center}
\FIG{8}{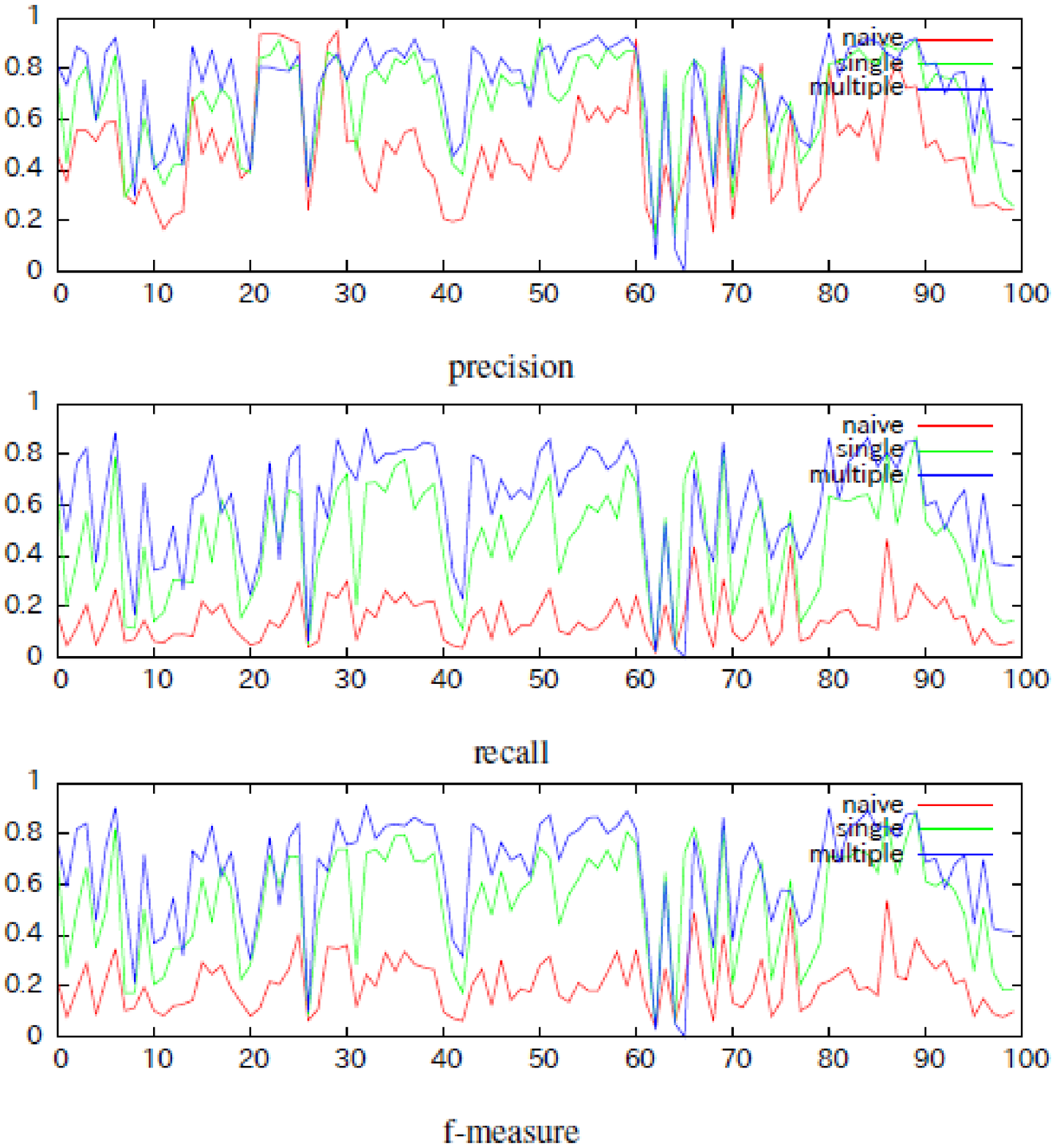}{}
\caption{
 Map-matching quality obtained in each map-matching task (``uncertain map" scenario).
}\label{fig:D}
\end{center}
\vspace*{-5mm}
\end{figure}
}

\newcommand{\figE}{
\begin{figure}[t]
\begin{center}
\FIG{8}{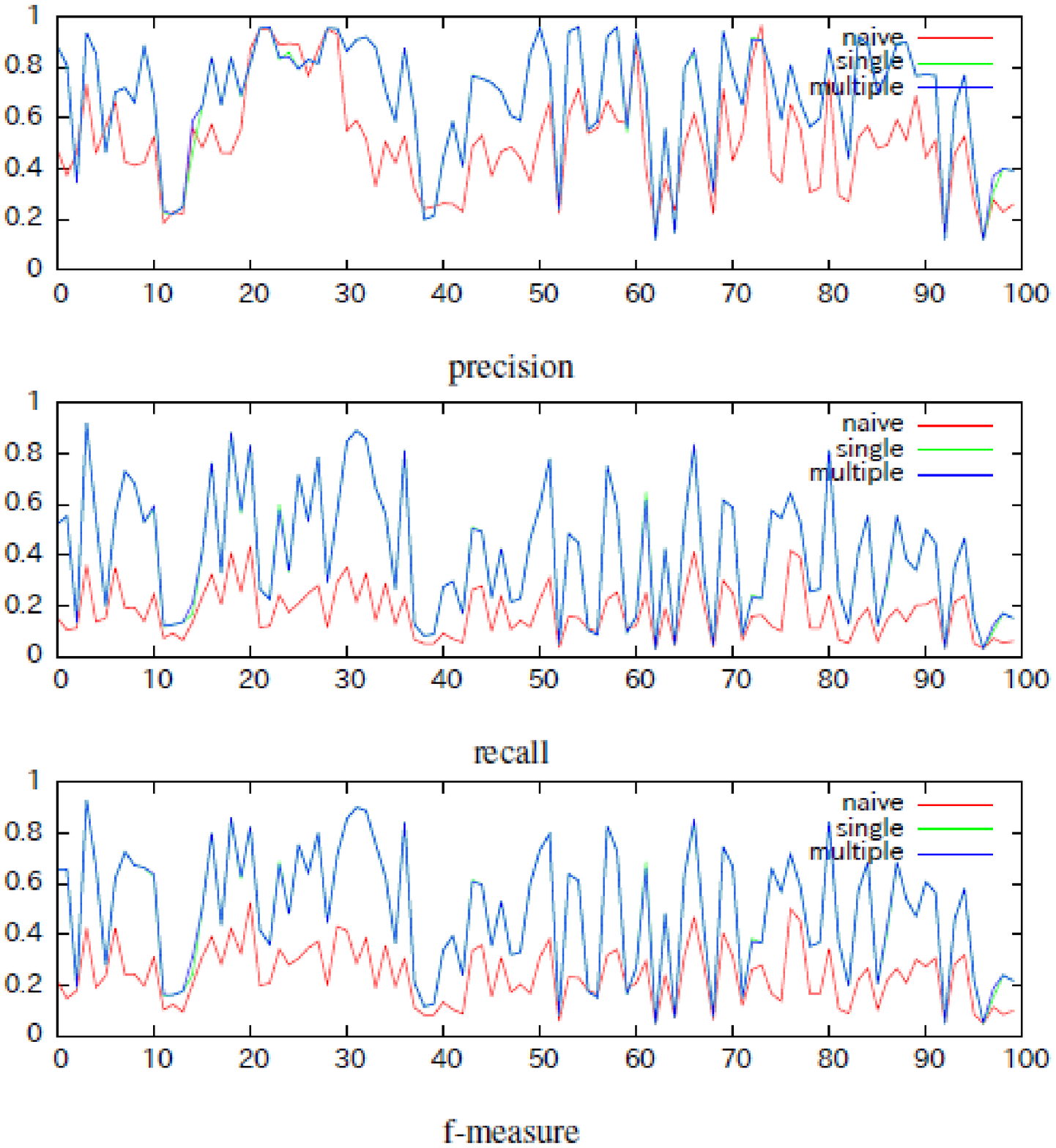}{}
\caption{
Map-matching quality obtained in each map-matching task (``precise map" scenario).
}\label{fig:E}
\end{center}
\end{figure}
}

\newcommand{\figH}{
\begin{figure}[t]
  \begin{center}
\FIG{8}{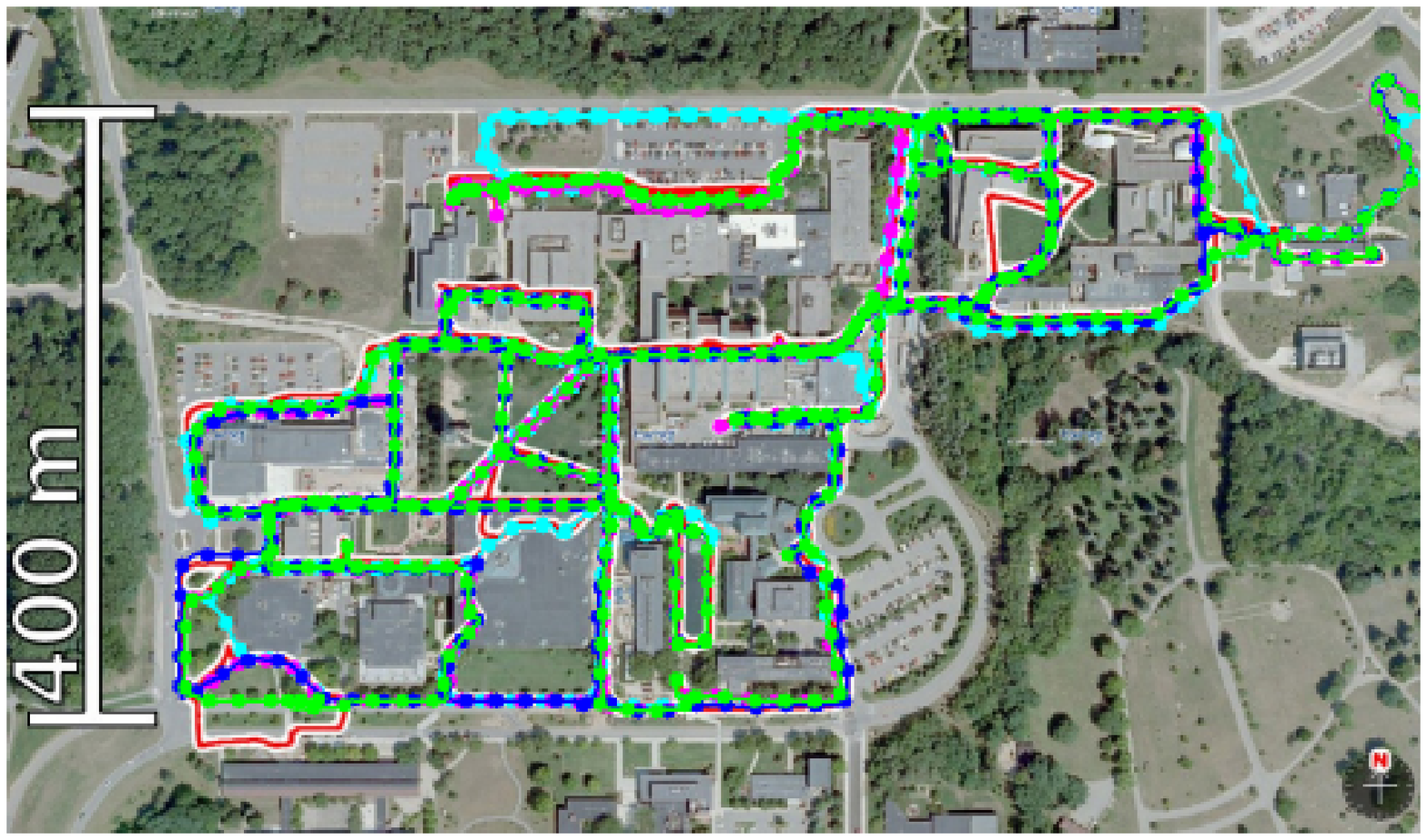}{}
\caption{
Experimental environment. The trajectories of the four datasets, ``2012/01/22 (01)," ``2012/03/31 (03)," ``2012/08/04 (08)," and ``2012/11/17 (11)," used in our experiments are visualized in green, purple, blue, and light-blue curves and overlaid on the bird's eye view imagery obtained from the NCLT dataset \cite{nclt}.
}\label{fig:H}
\end{center}
\vspace*{-5mm}
\end{figure}
}

\newcommand{\figI}{
\begin{figure}[t]
  \begin{center}
\FIG{8}{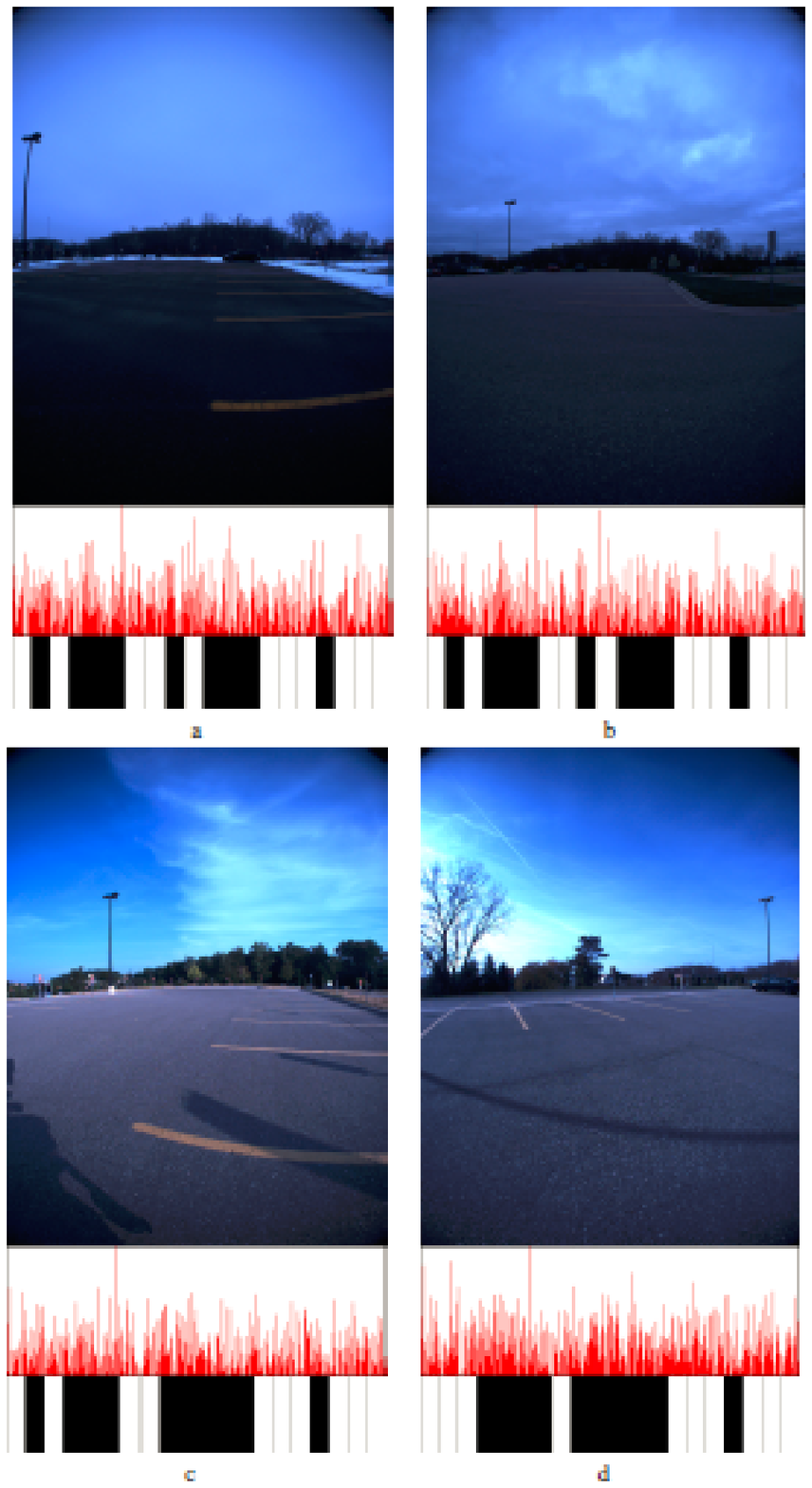}{}
\caption{
Examples of DCNN features. Four different visual images (200-th image in the dataset ``01," ``03," ``08," and ``11"), being explained by a 4,096-dim DCNN feature are shown. Each DCNN feature is further encoded to a 20-bit binary code that is visualized by a barcode.
}\label{fig:I}
\end{center}
\vspace*{-5mm}
\end{figure}
}

\newcommand{\figJ}{
\begin{figure}[t]
  \begin{center}
\FIG{8}{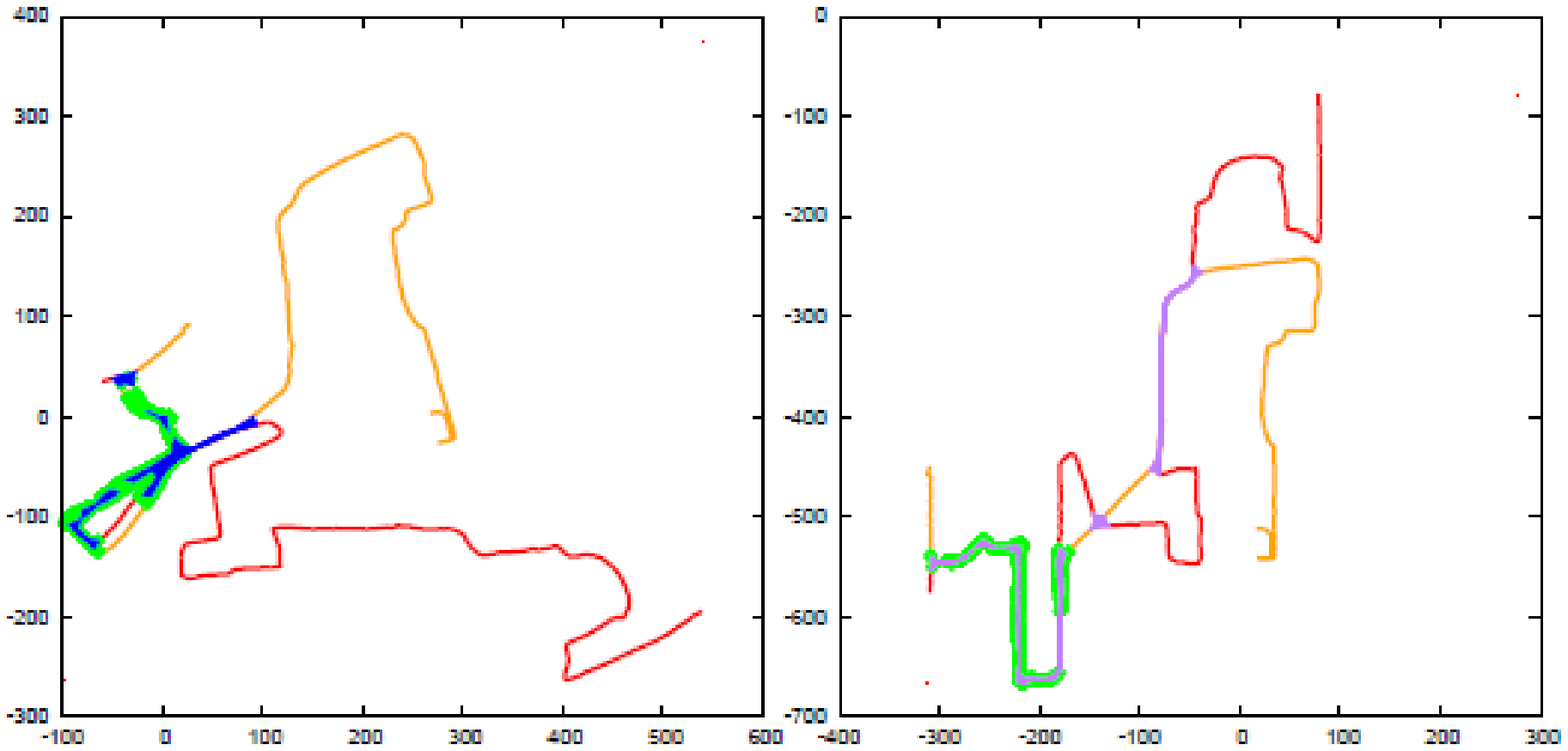}{}
\caption{
Map-matching task: Left: estimation; Right: ground-truth. Query and reference trajectories are shown in orange and red curves, respectively. Three types of correspondences, true positives (TP), false positives (FP), and false negatives (FN), are respectively indicated in green, blue, and purple lines. It should be noted that even with the discriminative DCNN features, matching two images with opposite viewing directions often fails, as shown in these FN examples.
}\label{fig:J}
\end{center}
\vspace*{-5mm}
\end{figure}
}

\newcommand{\tabC}{
\begin{table*}[t]
\begin{center}
\begin{minipage}[b]{5.5cm}
\begin{center}
\begin{tabular}{l|r|r|r|r|} \hline
top-X  & 1 & 2 & 5 & 10 \\ \hline\hline
naive  & 0.54 & 0.51 & 0.51 & 0.49 \\ 
single  & 0.72 & 0.71 & 0.69 & 0.67 \\ 
multiple  & 0.73 & 0.73 & 0.73 & 0.73 \\ 
\hline
\end{tabular}
precision
\end{center}
\end{minipage}
\begin{minipage}[b]{5.5cm}
\begin{center}
\begin{tabular}{l|r|r|r|r|} \hline
top-X  & 1 & 2 & 5 & 10 \\ \hline\hline
naive  & 0.18 & 0.16 & 0.16 & 0.15 \\ 
single  & 0.57 & 0.55 & 0.51 & 0.46 \\ 
multiple  & 0.63 & 0.62 & 0.62 & 0.61 \\ 
\hline
\end{tabular}
recall
\end{center}
\end{minipage}
\begin{minipage}[b]{5.5cm}
\begin{center}
\begin{tabular}{l|r|r|r|r|} \hline
top-X  & 1 & 2 & 5 & 10 \\ \hline\hline
naive  & 0.25 & 0.22 & 0.22 & 0.21 \\ 
single  & 0.63 & 0.61 & 0.58 & 0.53 \\ 
multiple  & 0.67 & 0.66 & 0.66 & 0.66 \\ 
\hline
\end{tabular}
f-measure
\end{center}
\end{minipage}
\caption{Average map-matching quality for top-$X$ ranked hypotheses (``uncertain map" scenario).}\label{tab:C}
\end{center}
\end{table*}
}

\newcommand{\tabD}{
\begin{table*}[t]
\begin{center}
\begin{minipage}[b]{5.5cm}
\begin{center}
\begin{tabular}{l|r|r|r|r|} \hline
top-X  & 1 & 2 & 5 & 10 \\ \hline\hline
naive  & 0.54 & 0.52 & 0.51 & 0.5 \\ 
single  & 0.68 & 0.68 & 0.68 & 0.68 \\ 
multiple  & 0.68 & 0.68 & 0.68 & 0.68 \\ 
\hline
\end{tabular}
precision
\end{center}
\end{minipage}
\begin{minipage}[b]{5.5cm}
\begin{center}
\begin{tabular}{l|r|r|r|r|} \hline
top-X  & 1 & 2 & 5 & 10 \\ \hline\hline
naive  & 0.19 & 0.18 & 0.17 & 0.18 \\ 
single  & 0.41 & 0.41 & 0.41 & 0.41 \\ 
multiple  & 0.41 & 0.41 & 0.41 & 0.41 \\ 
\hline
\end{tabular}
recall
\end{center}
\end{minipage}
\begin{minipage}[b]{5.5cm}
\begin{center}
\begin{tabular}{l|r|r|r|r|} \hline
top-X  & 1 & 2 & 5 & 10 \\ \hline\hline
naive  & 0.25 & 0.25 & 0.23 & 0.24 \\ 
single  & 0.49 & 0.49 & 0.49 & 0.49 \\ 
multiple  & 0.49 & 0.49 & 0.49 & 0.49 \\ 
\hline
\end{tabular}
f-measure
\end{center}
\end{minipage}
\caption{Average map-matching quality for top-$X$ ranked hypotheses (``precise map" scenario).}\label{tab:D}
\end{center}
\end{table*}
}

\author{Tanaka Kanji
\thanks{Our work has been supported in part by 
JSPS KAKENHI 
Grant-in-Aid for Young Scientists (B) 23700229,
and for Scientific Research (C) 26330297 
}
\thanks{K. Tanaka is with Faculty of Engineering, University of Fukui, Japan.
{\tt\small tnkknj@u-fukui.ac.jp}}%
\vspace*{-5mm}}

\begin{document}

\title{\LARGE \bf
Deformable Map Matching for Uncertain Loop-Less Maps
}

\maketitle

\author{}

\begin{abstract}
In the classical context of robotic mapping and localization, map matching is typically defined as the task of finding a rigid transformation (i.e., 3DOF rotation/translation on the 2D moving plane) that aligns the query and reference maps built by mobile robots. This definition is valid in loop-rich trajectories that enable a mapper robot to close many loops, for which precise maps can be assumed. The same cannot be said about the newly emerging autonomous navigation and driving systems, which typically operate in loop-less trajectories that have no large loop (e.g., straight paths). In this paper, we propose a solution that overcomes this limitation by merging the two maps. Our study is motivated by the observation that even when there is no large loop in either the query or reference map, many loops can often be obtained in the merged map. We add two new aspects to map matching: (1) image retrieval with discriminative deep convolutional neural network (DCNN) features, which efficiently generates a small number of good initial alignment hypotheses; and (2) map merge, which jointly deforms the two maps to minimize differences in shape between them. To realize practical computation time, we also present a preemption scheme that avoids excessive evaluation of useless map-matching hypotheses. To verify our approach experimentally, we created a novel collection of uncertain loop-less maps by utilizing the recently published North Campus Long-Term (NCLT) dataset and its ground-truth GPS data. The results obtained using these map collections confirm that our approach improves on previous map-matching approaches.
\end{abstract}

\section{Introduction}

Map matching is a fundamental problem for robotic mapping and localization. Given query and reference maps consisting of a sequence of visual images and odometry measurements, the goal is to find a set of corresponding robot poses (or viewpoints) between the two maps that are maximally consistent with the pairwise constraints from odometry and image matching (or loop closure \cite{ProbabilisticsRobotics}).

In previous studies, map matching is typically defined as the task of finding a rigid transformation (i.e., 3DOF rotation/translation on the 2D moving plane) that aligns the query and reference maps \cite{neira2003linear}. This definition is valid in loop-rich trajectories that enable a mapper robot to close many loops and build the maps precisely prior to the map matching. However, the same cannot be said about the newly emerging autonomous navigation and driving systems, which typically operate in loop-less trajectories that have no large loop (e.g., straight paths). Existing methods are not sufficiently robust against deformation of maps caused by a robot's odometry noises and map-building errors that are common in loop-less trajectories. They may assign low scores not only to wrong alignment hypotheses but also to correct hypotheses with map deformation.

In this paper, we overcome this limitation by merging the two maps (Fig. \ref{fig:A}). Our study is motivated by the fact that even when there is no large loop in either the query or reference map, many loops can often be obtained in the merged map. We add two new aspects to map matching: (1) image retrieval with discriminative deep convolutional neural network (DCNN) features, which efficiently generates a small number of good initial alignment hypotheses; and (2) map deformation, which jointly deforms the merged map to minimize shape differences between them.

\figA

The image retrieval approach is inspired by the recent success of visual features from DCNNs in visual place recognition (VPR) \cite{alexnet}. VPR uses a robot's visual image as a query input to search over a collection of reference images to locate a relevant reference image that is viewed from the nearest neighbor viewpoint to the query image's viewpoint. Recently, it has been found that the intermediate responses of a DCNN can be viewed as a discriminative feature for image retrieval. The current study is based on our recently developed DCNN-based image retrieval and re-ranking system \cite{iros16a}, in which a discriminativity-preserving dimension reduction by PCA \cite{lcd5} and compact binary codes \cite{IkedaICRA10} are employed for efficient visual search.

The map deformation framework is partially inspired by the state-of-the-art map-building techniques of graph SLAM \cite{ProbabilisticsRobotics}. The basic idea is to merge the two maps by connecting the pose pair of an image correspondence provided by the image retrieval system, and then to perform map deformation via graph SLAM using additional image correspondences as loop closure constraints. Theoretically, our deformation problem is solved by applying graph SLAM for the given odometry measurements and all possible combinations of loop closure constraints. However, such a brute-force procedure requires an infeasible amount of computation time, as there are typically many false positive matches even when the discriminative DCNN features are used. Further, the overall cost is proportional to the number of possible combinations of image correspondences.

Therefore, to realize practical computation time, we further employ an iterative preemption scheme. In the field of hypothesize-and-verify algorithms, preemption schemes are used to avoid excessive scoring of useless hypotheses by choosing between hypotheses given all the previous hypotheses and scoring results \cite{raguram2013usac}. We adopt the concept of an iterative preemption scheme to generate hypotheses. Instead of generating an intractable number of possible map hypotheses, our iterative algorithm selects a small number of top-ranked hypotheses at a time and uses each as a seed to generate the next generation of new hypotheses.

In the experiments conducted, we created a collection of loop-less maps that have no large loop, by utilizing recently published North Campus Long-Term (NCLT) image collections \cite{nclt} (Fig. \ref{fig:H}) obtained using the front directed camera (Cam\#5) of a vehicle-mounted Ladybug3 omnidirectional camera and its ground-truth GPS data. These map collections were used to compare our approach with previous map-matching approaches, which operate under the assumption of rigid transformation. For fair comparison, we developed and employed an autonomous dataset creation procedure that utilized the ground-truth GPS data. The results obtained indicate that our algorithm improves on previous map-matching    methods.

\section{Relation to Previous Work}

The primary contribution of this paper is a proposed practical method that overcomes the challenging map-matching problem faced by the newly emerging autonomous navigation and driving systems.

Map matching can be considered as a class of visual place recognition (VPR) tasks that have been intensively studied in the field of robot vision \cite{survey16vpr}. It is more challenging than the alternative popular loop closure detection class \cite{comparison09lcd}, which relies on the availability of an initial rough estimate of the robot pose. It is also more complex than the visual image retrieval class \cite{fabmap2}, in which the goal is to find a single image-level correspondence, rather than to find the set of all the correspondences between the maps.

Various types of map-matching approaches, ranging from visual features to hypothesize-and-verify algorithms, have been presented in the literature. However, most existing approaches assume that query and reference maps are precisely built {\it prior to} the map-matching task. In contrast, we focus on the more challenging and computationally demanding task of map matching with {\it posterior} map building (i.e., map deformation).

Our approach can be considered novel based on its incorporation of two independent recent developments in the robot vision community.

The first novelty lies in the use of the recently published NCLT dataset and its ground-truth GPS data \cite{nclt} to create a novel challenging dataset of {\it loop-less} trajectories. We employ the NCLT dataset to analyze overlap between trajectories and extract sub-trajectories that have no large loop. To the best of our knowledge, this is the first study in which an algorithm for automatic creation of loop-less trajectories is developed and employed in the above context.

The next novelty is the use of discriminative DCNN features that provide a small set of good initial hypotheses and enable a computationally tractable multiple-hypotheses approach to be developed. In particular, the underlying foundation of our algorithm is our recently developed DCNN-based image retrieval and re-ranking method \cite{iros16a}, which is successful in VPR tasks.

Furthermore, our approach is general and can be combined with various possible combinations of VPR algorithms, including visual features (e.g., DCNN features from convolutional layers), visual search (e.g., locality sensitive hashing), similarity metric (e.g., distance learning across domains), and map merging (e.g., joint optimization of multiple maps).

\section{Problem Formulation}\label{sec:prob}

\figH

Let us denote the pose sequences in the query and reference maps as $l_1^q$, $\cdots$, $l_Q^q$ and $l_1^r$, $\cdots$, $l_R^r$. Consequently, our goal is to find the correspondence between each $i$-th pose $i \in [1, Q]$ on the query map and its counterpart $r_i\in [0, R]$ on the reference map, which are maximally consistent with the pairwise constraints from odometry and image correspondences. Here, $r_i=0$ is prepared for the event where the $i$-th pose does not correspond to any reference pose.

In the classical formulation of map matching with rigid transformation (i.e., 3DOF rotation/translation), the quality of a map-matching hypothesis was measured in terms of the distance between an estimate and the ground-truth in the 3D transformation space. In contrast, the dimensionality of our solution space is proportional to the length of the robot's trajectory or the pose sequence and tends to be very large. Therefore, we present a map-matching quality measure extension that reflects a natural intuition about map-matching quality.

\figJ

More formally, we measure the quality of a map-matching hypothesis by comparing the pose correspondences on its merged map with those of the ground-truth (Fig. \ref{fig:J}). Given a map-matching hypothesis, each correspondence is first classified into true positive (TP), false positive (FP), or false negative (FN), according to whether it is consistent with both the hypothesis and the ground-truth, consistent with the hypothesis but not with the ground-truth, or not consistent with the hypothesis but consistent with the ground-truth. Here, a pose correspondence is considered consistent with a hypothesis or the ground-truth trajectory if and only if the Euclidean distance between the pose pair is smaller than a preset threshold of 10 m. Let $N^{TP}$ denote the number of TPs, $N^{FP}$ denote the number of FPs, and $N^{FN}$ denote the number of FNs. Based on the terminology, we can measure the hypothesis's quality by precision $s^p=N^{TP}/(N^{TP}+N^{FP})$, recall $s^r=N^{TP}/(N^{TP}+N^{FN})$, and f-measure $s=2s^ps^r/(s^p+s^r)$. Fig. \ref{fig:J} visualizes a typical example of TP, FP, and FN correspondences in a hypothesis and the ground-truth. It should also be noted that even with the discriminative DCNN features, matching two images with opposite viewing directions often fails, as shown in the figure.

We can now define map-matching performance based on the map-matching quality measure. Recalling that a map- matching algorithm outputs a ranked list of map-matching hypotheses, map-matching performance can be evaluated by averaging the quality values of the top-$X$ ranked map hypotheses (e.g., $X = 10$), where $X$ is a preset hyperparameter. More formally, performance is evaluated using three values: $\bar{s^p_X}=\sum_{i=1}^X s_i^p$, $\bar{s^r_X}=\sum_{i=1}^X s_i^r$, and $\bar{s_X}=\sum_{i=1}^X s_i$.

\section{Approach}

The proposed map-matching framework consists of two distinct stages: retrieval and re-ranking. The former stage retrieves the reference map using images in the query map as input, and outputs a ranked list of initial map alignment hypotheses (i.e., translation/rotation). The latter (re-ranking) stage first connects the two maps at one of the corresponding pose pairs provided by the image retrieval system, performs iterative map deformation on the merged map, and then re-ranks the hypotheses on the basis of consistency between the deformed map pairs. These stages are detailed in the ensuing subsections.

\figI

\figC


\subsection{Retrieval}\label{sec:retrieval}

The image retrieval system encodes the image to a DCNN feature representation, as in \cite{alexnet}. First, it extracts a 4,096 dimensional DCNN feature from the given image. We use the sixth layer of DCNN because it has been proven to produce effective features with excellent descriptive power in previous studies \cite{lcd5}. We then perform PCA compression to obtain 128 dimensional features. This strategy is supported by the recent findings in \cite{lcd5}, in which PCA compression provided excellent short codes with 128 short vectors that generate state-of-the-art accuracy on several recognition tasks. However, direct use of DCNN features for image retrieval is computationally demanding, as it requires many-to-many comparisons of high-dimensional DCNN features between the query and reference images.

To address this concern, we perform compact binary images encoding and hash table indexing that enable fast image comparison. We encode query and library features to 20-bit binary codes using the compact projection technique \cite{tomomi2011incremental} and then consider those images with Hamming distances to the query code shorter than threshold $N_b=1$ as candidates of the image correspondences. The L2 distance in the high-dimensional DCNN features between the image pair of each candidate is then computed and the top $N_r (= 10)$ elements with the smallest L2 distance are accepted as an image correspondence. Note that the time cost for the comparison is proportional to the length of the candidate list, which is equal to the number of items stored in the corresponding bucket in the hash table. To achieve real-time computation, we ignore those entries in the buckets that store a greater number of items than a preset threshold of 100. As explained in our previous study \cite{icra08kanji}, this strategy not only contributes to reduction of the time cost for comparison, it also prevents wastage of the memory space by storing non-distinctive visual features. Fig. \ref{fig:I} shows several examples of input images, DCNN features, and binary codes.

The image retrieval system provides many-to-many image correspondences between the query and reference images. $N_r$ correspondences need to be selected (as the initial map alignment hypotheses) from the many-to-many correspondences. In this study, we simply selected the $N_r$ correspondences with the smallest L2 distances over all the image correspondences on which the L2 distance is computed. We empirically found that this simple strategy works well and provides adequate alignment hypotheses in practice.

Computation of the pose correspondences for each initial hypothesis is straightforward. First, all the robot poses on the query map are rotated/translated by the hypothesized transformation (i.e., rotation/translation). Next, the Euclidean distance between each of the transformed robot poses and its nearest neighbor pose in the reference map is computed and, if the Euclidean distance is smaller than the 10 m preset threshold, the pose pair is considered {\it consistent} against the map hypothesis. The reference pose with the lowest Euclidean distance is assigned to each query pose if the correspondence is consistent; otherwise, no reference pose is assigned (i.e., $s_t = 0$).

\subsection{Re-ranking}\label{sec:reranking}

The re-ranking stage jointly deforms both the query and reference maps to minimize differences in shape between them, and re-ranks the hypotheses based on consistency between the deformed map pairs. The basic procedure consists of two steps: (1) map merge, and (2) map deformation.

The first step connects the two maps at one of the $K$ pose pairs from the $K$ image correspondences. At the same time, the relative pose of the query pose with respect to the reference pose is computed and stored as the 3DOF constraint for future use by graph SLAM.

The second step iterates between addition of a new loop constraint to the merged map and graph SLAM using all the previous constraints obtained thus far. However, the problem of how to select a new constraint at each iteration exists. Intuitively, the next constraint, $z^{(2)}$, should be {\it inconsistent} with the previous map hypothesis, $h^{(1)}$, because we wish to obtain a new map hypothesis, $h^{(2)}$, that is dissimilar to the existing hypothesis, $h^{(1)}$. In general, the $i$-th constraint $z^i$$=(q^{(i)},{r}^{(i)})$ ($q\in[1,Q]$, $r\in[1,R]$) should be inconsistent with the trajectory hypothesis $h^{(i-1)}$. To implement this idea, we select the next constraint $z^{(i)}$ from those constraints $\{(q,r)\}$ in which the Euclidean distance between the pose pair exceeds a predefined threshold, $T_p$. In this study, $T_p$ was empirically set to 1 m. Fig. \ref{fig:C} shows an example of sequential hypothesis generation obtained using the proposed iterative algorithm.


In practice, computational resources are always limited and, as the number of hypotheses increases, smart use of computational resources becomes very important. Preemption is used to avoid excessive evaluation of useless hypotheses. This is accomplished by first generating $K = 10$ hypotheses via map merging from $K$ image correspondences. All the hypotheses are then scored by counting the number of consistent image correspondences. On the basis of these scores, the best $K' = 10$ hypotheses are selected and each is used in the map merging procedure to generate a new hypothesis. The process is then repeated, generating $K'$ new hypotheses from the best $K'$ hypotheses obtained up to that point. The above algorithm generates $K + K'M$ map hypotheses after $M = 10$ iterations, which is a reasonably low cost in our experiments.

\section{Experiments}

To evaluate our methods, we created a new map collection comprising uncertain loop-less maps. Using the NCLT dataset and its ground-truth GPS data \cite{nclt}, we obtained a number of samples of short trajectories with corresponding ground-truth GPS data that have no large loop and a visual image sequence. This enabled us to evaluate the performance of the map-matching algorithm proposed in Section \ref{sec:prob}. More formally, a trajectory is regarded as having no large loop unless there exists at least one pair of robot poses on the reference map's trajectory that (1) are farther from each other than 100 m in terms of travel distance and (2) both have image correspondences. We used four datasets from the NCLT ``2012/01/22," ``2012/03/31," ``2012/08/04," and ``2012/11/17" collection, corresponding to four different seasons of the same environment, as shown in Fig. \ref{fig:H}, respectively denoted as ``01," ``03," ``08," and ``11." We down-sampled the poses to obtain a lower temporal resolution of approximately one pose per meter travel distance, simulated odometry noises with $1\%$ standard deviations in translation and rotation, and obtained four image collections with sizes 2618, 2632, 2410, and 2450. We then considered 12 possible combinations of (query, reference) pairs: (``01", \{``03", ``08", ``11"\}), (``03", \{``01", ``08", ``11"\}), and (``08", \{``01", ``03", ``11"\}), (``11", \{``01", ``03", ``08"\}). For each dataset pair, we chose 10 pairs of query and reference trajectories by considering the fact that each trajectory pair needs to have view overlap, otherwise the map-matching task has no valid solution. To meet the view overlap requirement, we explicitly defined view overlap as the maximum travel distance between all possible pairs of reference poses that have image correspondences with some query images. Then, we sorted all the trajectory pairs by view overlap and selected the 10 trajectory pairs with the largest view overlap for our dataset. We found that two of the dataset pairs, (``08", ``03") and (``08", ``11"), had no such trajectory pair with a view overlap, and thus used the other 10 dataset pairs to create $10\times 10$ trajectory pairs for the experiments.

\figD

\tabC

\figB

\tabD

We considered two different map-matching task scenarios. In one scenario, termed the ``uncertain map" scenario, both the query and reference maps were uncertain and loop-less. In the other scenario, termed the ``precise map" scenario, only the query map was uncertain.

For the ``uncertain map" scenario, we compared three different map-matching algorithms: ``naive," ``single," and ``multiple." The ``naive" algorithm was the previous map-matching approach that assumed rigid transformation and scores each hypothesis by L2 distance between DCNN features from the single pair of images that are also used by the initial map alignment. The ``single" algorithm first generates a set of $K = 10$ initial hypotheses and then, for each initial hypothesis, it generates a new hypothesis by adding a single new constraint and performing graph SLAM using the two constraints. The ``multiple" algorithm was the proposed algorithm that, as explained in Section \ref{sec:reranking}, iterates between addition of a new constraint and graph SLAM. The two parameters, $K'$ and $M$, i.e., the number of top-ranked hypotheses used to generate the next generation of hypotheses per iteration and the number of iterations, were set to $K'=10$ and $M=10$.

Fig. \ref{fig:D} graphically reports the map-matching quality for all the 100 map-matching tasks considered here. It can be seen that the proposed ``multiple" algorithm clearly outperformed the other two algorithms, and ``single" algorithm was better than ``naive" algorithm. By closing many loops in the merged map, the proposed algorithm was able to achieve high accuracy in the merged map and also in its submaps (i.e., query/reference maps). As a result, it was able to obtain better pose correspondences with higher accuracy.

Fig. \ref{fig:B} gives map-matching examples for the three algorithms with the ground-truth. It can be seen that the proposed approach, deformable map matching, was effective for most of the cases. In addition, ``multiple" algorithm was better than ``single" algorithm because it effectively handled more complex cases, whereas the merged map had multiple large loops.

Table \ref{tab:C} summarizes the map-matching quality averaged over the 100 map-matching tasks for different settings of $X =$ $1$, $2$, $5$, and $10$. It is clear that the performance of the proposed approach is not sensitive to the choice of $X$.

Fig. \ref{fig:E} and Table \ref{tab:D} present the results for the alternative ``precise map" scenario, in which the reference map was precisely built prior to the map-matching task. In this case, we created the reference maps without adding the simulated odometry noises. It can be seen that the performance difference between ``single" and ``multiple" algorithms is not significant in this case, and also that ``single" and ``multiple" algorithms clearly outperformed the ``naive" algorithm. This is mainly because closing a single loop in the merged map is already sufficient to achieve high accuracy in the merged map, and we often did not need additional multiple loops to be closed. Thus, it can be concluded that the proposed map-matching approach is effective for uncertain loop-less maps, with a performance gain that tends to be significant when both the query and reference maps have no large loop.

\figE

\section{Conclusions}

In this paper, we proposed a practical solution called ``deformable map matching" that overcomes the challenging map-matching scenario of ``uncertain loop-less maps" confronting emerging autonomous navigation and driving systems. The proposed method overcomes the inability of existing methods to handle loop-less trajectories (e.g., straight paths) that have no large loop by merging the query and reference maps. Our study was motivated by the fact that even when no large loop exists in either the query or reference map, many loops can often be obtained in the merged map. We added two new aspects to map matching: (1) image retrieval with discriminative DCNN features, which efficiently generates a small number of good initial alignment hypotheses; and (2) map merge, which jointly deforms the merged map to minimize shape differences between the two maps. To realize practical computation time, we further utilized a preemption scheme that avoids excessive evaluation of useless map-matching hypotheses. Furthermore, we created a novel collection of uncertain loop-less maps by utilizing the recently published North Campus Long-Term (NCLT) dataset and its ground-truth GPS data. The experimental results obtained using these map collections verify that our algorithm improves on previous map-matching methods.

\bibliographystyle{IEEEtran}
\bibliography{mhv}

\end{document}